\documentclass[10pt,twocolumn,letterpaper]{article}

\usepackage{cvpr}              
\usepackage{multicol}
\usepackage{multirow}
\usepackage{utfsym}
\usepackage{bbding}
\usepackage{amssymb}

\definecolor{cvprblue}{rgb}{0.21,0.49,0.74}
\usepackage[pagebackref,breaklinks,colorlinks,citecolor=cvprblue]{hyperref}

\def\paperID{****} 
\def\confName{CVPR}
\def\confYear{2024}

\title{Catastrophic Overfitting: A Potential Blessing in Disguise}

\author{Mengnan Zhao, Lihe Zhang\thanks{Corresponding author}, Yuqiu Kong, Baocai Yin
}

\begin{document}
\maketitle
\begin{abstract}
Fast Adversarial Training (FAT) has gained increasing attention within the research community owing to its efficacy in improving adversarial robustness. 
Particularly noteworthy is the challenge posed by catastrophic overfitting (CO) in this field.  
Although existing FAT approaches have made strides in mitigating CO, the ascent of adversarial robustness occurs with a non-negligible decline in classification accuracy on clean samples. 
To tackle this issue, we initially employ the feature activation differences between clean and adversarial examples to analyze the underlying causes of CO. 
Intriguingly, our findings reveal that CO can be attributed to the feature coverage induced by a few specific pathways. 
By intentionally manipulating feature activation differences in these pathways with well-designed regularization terms, we can effectively mitigate and induce CO, providing further evidence for this observation.
Notably, models trained stably with these terms exhibit superior performance compared to prior FAT work.
On this basis, we harness CO to achieve `attack obfuscation', aiming to bolster model performance.
Consequently, the models suffering from CO can attain optimal classification accuracy on both clean and adversarial data when adding random noise to inputs during evaluation.
We also validate their robustness against transferred adversarial examples and the necessity of inducing CO to improve robustness.
Hence, CO may not be a problem that has to be solved.
\end{abstract}
\section{Introduction}
\begin{figure}[t]
    \begin{center}
    \vspace{-2mm}
        \includegraphics[width=0.92\linewidth]{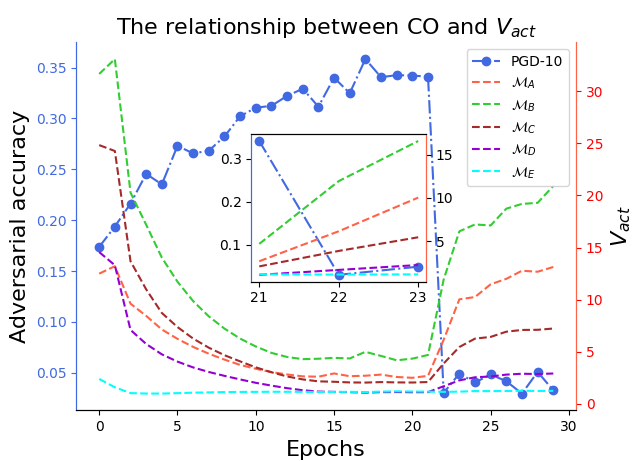}
    \end{center}
    \vspace{-7mm}
    \caption{Relationship between catastrophic overfitting and feature activation differences $V_{act}$ during adversarial training.
    We select CIFAR-10 \cite{krizhevsky2009learning} as the dataset and ResNet18 \cite{he2016deep} as the network $\mathcal{M}$. 
    Five activation nodes $\mathcal{M}_{\text{A}\sim\text{E}}$ are chosen from $\mathcal{M}$.
    Left y-axis: Model robustness against the PGD-10 attack;
    Right y-axis: $V_{act}$ on various activation nodes. 
    $V_{act}$ is quantified using $\mathcal{L}_2$ regularization between features of clean and adversarial examples.
    }
    \label{fig1}
    \vspace{-4mm}
\end{figure}
In recent years, the realm of deep learning has witnessed remarkable advancements \cite{zuo2022deep,mousavi2022deep,pereira2022sleap}.
However, the widespread adoption of deep neural networks \cite{borisov2022deep,salahuddin2022transparency} has inevitably prompted researchers to examine their limitations \cite{chou2023backdoor,wei2022noise,chen2022effective}, particularly their vulnerability to adversarial attacks \cite{cao2022advdo,gu2022segpgd,zhong2022shadows}. 
For improving model robustness, adversarial training \cite{mo2022adversarial,jia2022adversarial,wu2022towards} serves as a typical technique that incorporates perturbed samples into the training dataset. 
Nevertheless, conventional adversarial training approaches \cite{xiao2022stability,jin2022enhancing,guzman2022cross} bear the burden of time-intensive processes, 
\ie, generating adversarial training data using multi-step attacks 
such as the projected gradient descent (PGD) \cite{madry2017towards}. 
Therefore, researchers have progressively shifted their attention towards fast adversarial training (FAT) \cite{li2022subspace,zhang2022revisiting,jia2022boosting}.

FAT methods commonly employ a one-step, fast gradient-sign method (FGSM) \cite{goodfellow2014explaining} to generate adversarial training data. 
While these approaches are more time-efficient than PGD-based adversarial training, they often face the challenge of catastrophic overfitting (CO) \cite{de2022make,rice2020overfitting}, 
\ie, the classification accuracy of trained models for adversarial examples encounters a sudden decrease. 
This challenge becomes increasingly prevalent as the perturbation budget $\xi$ is expanded during adversarial training \cite{zhao2023fast}. 
To solve it, researchers are dedicated to enhancing the diversity of adversarial perturbations \cite{Wong2020,huang2022fast} and introducing additional regularization terms \cite{Andriushchenko2020,Jiag2022prior}. 
Even though these advanced techniques exhibit notable progress in addressing the stability concern,
their performance on clean samples significantly declines. 
For instance, ResNet18 \cite{he2016deep} can achieve an impressive 94\% classification accuracy on clean CIFAR-10 \cite{krizhevsky2009learning} samples, but the accuracy drops to 64\% when $\xi$ = 16/255 during adversarial training, as shown in \cite{zhao2023fast}. This prompts us to explore CO in depth. 

He et al. \cite{he2023investigating} think that the self-fitting results in CO
and utilize a high-variance channel masking approach to eliminate the attack information. 
However, the channels responsible for learning data information can also exhibit high variance, as demonstrated in our experiments. 
Therefore, we introduce a more effective channel-selecting method -- feature activation differences between clean and adversarial examples, to provide
a comprehensive explanation for CO.
As depicted in Fig. \ref{fig1}, CO is accompanied by salient feature activation differences. 
By deliberately suppressing or enlarging feature activation differences of several specific channels using well-designed regularization terms, we can effectively address or induce CO, respectively.
This actually further verifies the relationship between CO and salient feature differences.
%
%
Notably, under stable adversarial training, the proposed regularization terms approach zero and are thus insensitive to the choice of hyperparameters.


On this basis,  we try to improve the adversarial robustness of CO-affected models by leveraging CO, while maintaining their performance on clean samples.
This can be realized by introducing processing techniques that possess threefold capacities:
1) preserve valuable data information; 
2) disrupt potential attack information embedded in adversarial examples; 
3) have timeliness.
As a result, adding random noise to model inputs during evaluation is a simple yet effective manner.
It helps CO-affected models realize optimal classification accuracy for both clean and adversarial data, and does not improve the performance of models exclusively trained on clean samples. 
We explain this phenomenon from a novel perspective of ``attack obfuscation''.

The contributions are summarized as follows:
{\bf (1)} We systematically analyze the causes of CO by examining feature activation differences between clean and adversarial examples. 
These differences can more precisely identify the specific channels responsible for CO than previous studies;
{\bf (2)} We present novel regularization terms to avoid (or induce) CO by deliberately suppressing (or amplifying) feature activation differences in several selected channels.
Models trained stably with these terms achieve better or comparable performance to state-of-the-art FAT techniques;
{\bf (3)} We leverage CO to enhance model performance. By applying random noise to model inputs during evaluation, CO-affected models attain optimal classification accuracy for both clean and adversarial examples. We analyze this phenomenon from the perspective of attack obfuscation and give full experimental evidence. 

\section{Related work}
To improve the adversarial robustness of deep models, researchers commonly employ adversarial training techniques, \ie, enhancing the training dataset $D_{train}$ with adversarial perturbations ($\delta$). 
A typical adversarial training formula \cite{madry2017towards} tackles a min-max optimization problem,
\begin{equation}\label{re1}
\min_\theta \mathbb{E}_{(x, y)\sim D_{train}} \left[\max_{\delta\in[-\xi, \xi]} \mathcal{L}\left(\mathcal{M}(x+\delta), y\right) \right],
\end{equation}
where $\mathcal{M}(\cdot)$ denotes a fixed victim model parameterized by $\theta$. 
$\xi$ describes the maximum perturbation budget.
$\mathcal{L}(\cdot)$ is typically the cross-entropy loss. 
The internal maximization often yields un-targeted adversarial examples based on the original targets $y$ of inputs $x$.
An alternative adversarial training paradigm \cite{sriramanan2021towards,zhang2019theoretically} applies a regression constraint to clean samples
and introduces regularization terms to align the predictions for clean $x$ and adversarial examples $x+\delta$, 
\begin{equation}\label{re2}
\min_\theta \mathbb{E}_{(x, y)\sim D_{train}}\left[ \mathcal{L}\left(\mathcal{M}(x), y\right) + \|\mathcal{M}(x+\delta) - \mathcal{M}(x)\|_* \right],
\end{equation}
where $\|\cdot\|_*$ signifies a norm function. $\delta$ in both Eqs. (\ref{re1}) and (\ref{re2}) is usually generated by the 
gradient-based adversarial attacks \cite{yuan2021meta,wan2023average,papernot2016limitations}.
For instance, Goodfellow et al. \cite{goodfellow2014explaining} proposed a single-step gradient attack -- FGSM, 
\begin{equation}\label{re3}
    \delta = \xi\cdot \text{sign}(\nabla_{x}\mathcal{L}(\mathcal{M}(x), y)), 
\end{equation}
where $\text{sign}(\cdot)$ is the sign function. $\nabla_{x}\mathcal{L}(\cdot)$ denotes the gradient backward.
Similarly, Cheng et al. \cite{cheng2021fast} introduced the fast gradient non-sign method, which incorporates a scale factor $\zeta$ 
to control 
$\xi$,
$\zeta = \frac{\|\text{sign}(\delta)\|_2}{\|\delta\|_2}$.
Apart from single-step attacks, there are also multi-step attacks \cite{Kurakin2017,dong2018boosting,madry2017towards} that employ a small stride $\epsilon$ and restrict the $\xi$
at each step $t$,
\begin{equation}
    \begin{aligned}
    x^\prime_{t+1} = &\text{clip}_\xi(x^\prime_{t}+ \epsilon\cdot \text{sign}(\nabla_{x^\prime_{t}}\mathcal{L}(\mathcal{M}(x^\prime_{t}), y))),\\
    &\delta = x^\prime_{final} - x.
    \end{aligned}
\end{equation}

\begin{figure*}[t]
    \begin{center}
        \includegraphics[width=0.83\linewidth]{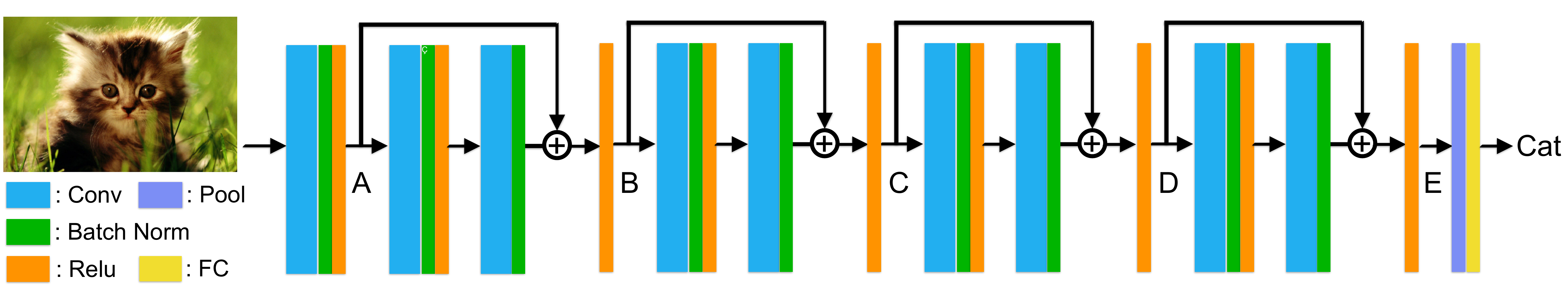}
    \end{center}
    \vspace{-6mm}
    \caption{Overall architecture and the placement of activation nodes. We select five activation nodes in the network ResNet18, and each node locates after a ReLU function.}
    \label{fig2}
    \vspace{-4mm}
\end{figure*}
\begin{figure}[t]
    \begin{center}
        
        \includegraphics[width=1\linewidth]{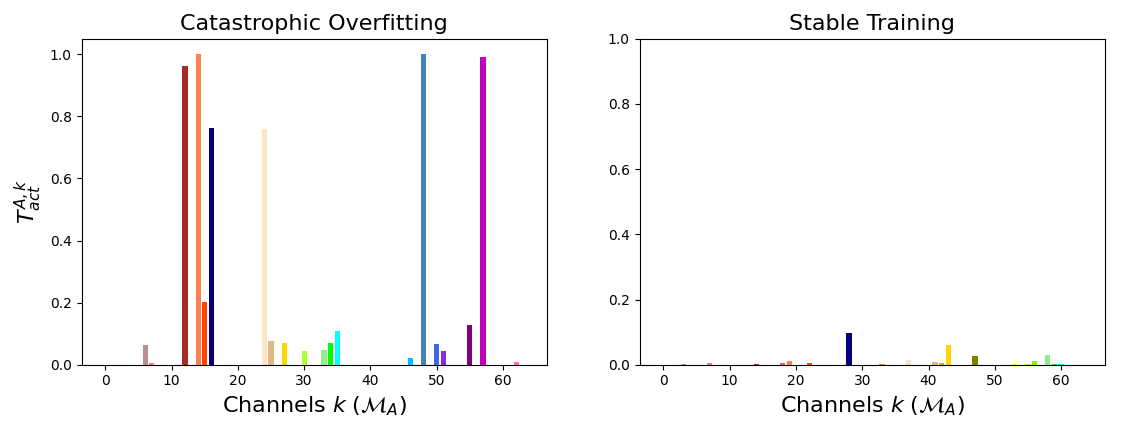}
    \end{center}
    \vspace{-5mm}
    \caption{
    Statistical analyses of channel-specific feature activation differences at the activation node $\mathcal{M}_\text{A}$. }
    \vspace{-4mm}
    \label{fig3}
\end{figure}
Based on the types of adversarial attacks, adversarial training techniques can be broadly categorized into two kinds, \ie, fast adversarial training methods \cite{jia2022boosting,park2021reliably,shafahi2019adversarial} utilizing single-step attacks like FGSM and conventional methods that rely on multi-step attacks \cite{jin2022enhancing,wu2022towards,tramer2019adversarial} such as PGD.
Compared to PGD-based methods, FGSM-based approaches require less time but are susceptible to CO.
To mitigate CO, various techniques have been introduced, such as FGSM-RS \cite{Wong2020}, GradAlign \cite{niu2022fast}, NuAT \cite{sriramanan2021towards}, ZeroGrad \cite{Golgooni2021}, and ConvergeSmooth \cite{zhao2023fast}.
However, the enhancement of adversarial robustness brought by these methods often decreases the classification accuracy on clean samples. 
Furthermore, increasing the perturbation budget during adversarial training will further reduce this accuracy.

Inspired by \cite{he2023investigating}, we revisit CO.
Surprisingly, we observe that by simply introducing random noise to inputs during evaluation, the CO-affected models can achieve excellent adversarial robustness while maintaining their classification accuracy on clean examples. 
This discovery challenges the conventional cognition and raises a question about whether CO is indeed a problem that requires to be solved.

\section{What causes catastrophic overfitting?}\label{sec3}
In this section, we investigate the causes of CO through experiments.
For a fair comparison, we follow \cite{Jiag2022prior,zhao2023fast} to realize the FAT baseline, FGSM-MEP \cite{Jiag2022prior}.
The perturbation budget during adversarial training ($\xi_T$) is set to 12/255 to ensure the occurrence of CO.
We expand the perturbation budget for evaluation ($\xi_E$) to 16/255 as the uncertainty perturbation degree of attacks.
The 
larger values of $\xi_E$ are not considered since they may change human predictions \cite{addepalli2022scaling}.

\subsection{Relationship between CO and $V_{act}$ }\label{sec3.1}
\begin{figure}[t]
    \begin{center}
        \vspace{-2mm}
        \includegraphics[width=1.\linewidth]{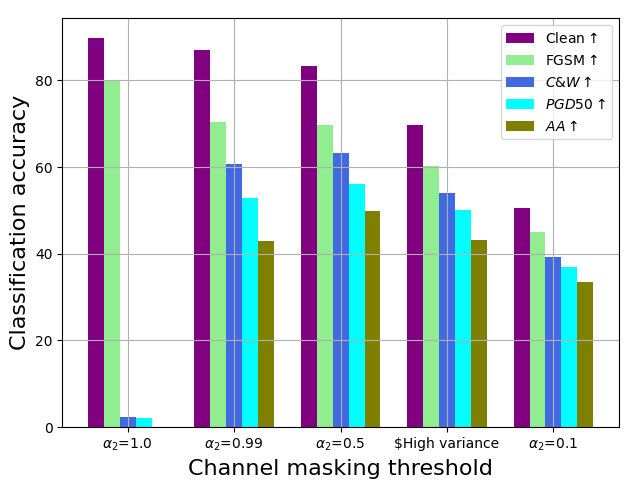}
    \end{center}
    \vspace{-6mm}
    \caption{Classification accuracy of models suffering from CO on clean and adversarial examples under various channel masking thresholds. `$\alpha_2$=1.0' indicates the original classification accuracy. `$\$$High variance' signifies using \cite{he2023investigating} to mask an equivalent number of channels as our method when `$\alpha_2$=0.5'.
    }
    \label{fig4}
    \vspace{-4mm}
\end{figure}

Prior work \cite{he2023investigating} believes that high-variance feature channels are responsible for CO. 
However, we find that masking these channels greatly affect the classification accuracy on clean samples, as shown in Fig. \ref{fig4}.
Therefore, we introduce the feature activation differences to more precisely determine the channels closely associated with CO.



Taking ResNet18 as an example, five activation nodes are selected, with each node located after a ReLU activation function. 
Their locations are visually depicted in Fig. \ref{fig2}.
Then, we quantify the feature activation differences
between the clean examples $x$ and adversarial examples $x+\delta$ 
at selected activation nodes by
\begin{equation}\label{eq1}
\centering
    \begin{aligned}
    V_{act}^i(x, x+\delta) = \frac{1}{\text{M}}\sum_{x\in D_{train}}\|\mathcal{M}_i(x) - \mathcal{M}_i(x+\delta)\|_2,
    \end{aligned}
\end{equation}
where $D_{train}$ is the training dataset and $\text{M}$ is the number of samples in $D_{train}$. 
$\mathcal{M}_i(x)$ is features of $x$ at the $i$-th activation node $\mathcal{M}_i$. $\|\cdot\|_2$ denotes the 2-norm function.

Fig. \ref{fig1} shows the adversarial robustness of models and 
feature activation differences $V_{act}^i$ 
per training epoch.
In phases of stable adversarial training, $V_{act}^{\text{A}\sim \text{C}}$ converge to a small value. 
However, they perform a sudden increase when models fall into CO.
Furthermore, experimental results on various datasets and networks in supplementary materials exhibit similar phenomena. 

{\bf 
Impact of $\delta$ to different channels.}\label{sec3.2}
Now we know that CO is accompanied by salient feature activation differences. 
Next, we investigate if all or only several specific pathways show such differences.
To determine this, we calculate channel-wise feature activation differences at $\mathcal{M}_\text{A}$ on the test dataset $D_{test}$. 
We also conduct experiments on other nodes and show results in supplementary materials.
\begin{equation}\label{eq2}
\centering
    \begin{aligned}
    &V_{act}^{{i},k}(x, x+\delta) = \|\mathcal{M}_{{i},k}(x) - \mathcal{M}_{{i},k}(x+\delta)\|_2^2,\\
    &T_{act}^{{i},k} = \tanh(\alpha\cdot \sum_{x\in D_{test}}\frac{V_{act}^{{i},k}(x, x+\delta)}{\text{N}\cdot \text{H}\cdot \text{W}}),
    \end{aligned}
\end{equation}
where $\mathcal{M}_{i}(x)\in\mathbb{R}^{\text{N}\times \text{H}\times \text{W}}$, $\mathcal{M}_{{i},k}$ denotes the $k$-th channel of $\mathcal{M}_{{i}}$, with $k\in [1,\text{N}]$. $\text{N}$, $\text{H}$ and $\text{W}$ correspond to the number of feature channels, the height and width of feature maps, respectively. 
$\|\cdot\|_2^2$ represents the squared 2-norm function.
$\alpha$ is set to 100 for improved visualization.

Fig. \ref{fig3} displays the channel-wise statistical information of $T_{act}^{\text{A},k}$. The observations are as follows:
1) For models that suffer from CO, while most channels show a small $T_{act}^{\text{A},k}$, 
a few channels exhibit significantly large values;
2) During stable adversarial training, $T_{act}^{\text{A},k}$ of all channels tends to approach zero.
Additionally, we find that the salient $T_{act}^{\text{A},k}$ consistently occurs at the same channels across different adversarial examples. 
These findings suggest that adversarial perturbations mainly influence several specific pathways.

{\bf Exploring the role of different channels.}\label{sec3.3}
Building upon the above experimental analyses,
we propose an empirical hypothesis: channels with salient feature activation differences 
are primarily utilized to learn attack information. 
To verify this, 
we compare the classification performance of various masked models on both clean and adversarial examples.
These models are obtained by masking diverse channels of a CO-affected model,
\begin{equation}\label{eq3}
\centering
    \begin{aligned}
    \mathcal{M}_{{i},k}(x+\delta)\leftarrow 0 \quad \text{if}\quad T_{act}^{{i},k} > \alpha_2.
    \end{aligned}
\end{equation}
Based on statistic results of Fig. \ref{fig3}, we set the masking threshold $\alpha_2$ to 1.0, 0.99, 0.5, and 0.1, respectively.
Furthermore, the adversarial examples are generated against the original model ($\alpha_2$ = 1.0)
using FGSM \cite{goodfellow2014explaining}, PGD \cite{madry2017towards}, C$\&$W \cite{carlini2017towards}, and Autoattack (AA) \cite{Francesco2020}.
They are directly used as inputs of masked models to evaluate.
Fig. \ref{fig4} shows the detailed evaluation results.
(1) Masking the channels 
of $T_{act}^{\text{A},k}>0.99$ 
can efficiently improve the model performance on adversarial examples while only slightly affecting the performance on clean samples. 
(2) This trend continues when reducing $\alpha_2$ to 0.5. 
(3) However, when $\alpha_2$ is set to 0.1,  the model performance on clean samples greatly decreases. 
These observations support our hypothesis.
Notably, model channels cannot be directly divided into data or adversarial branches.
This is because 
masking the pathways always influences the classification accuracy on clean samples.

Similarly, we provide the results of variance-based \cite{he2023investigating} channel masking in Fig. \ref{fig4}. Under an equal number of masked channels, the proposed activation difference shows superior prediction performance on both clean and adversarial data. 
This indicates the effectiveness of our approach in selecting pathways that extract attack information.

\begin{table}
	\centering
	\tabcolsep = 0.085cm
 \begin{center}
	\begin{tabular}{ccccc cccc} 
		\hline
		   Nodes&$\alpha_3$  &  \textit{Clean}&  \textit{FGSM}& \textit{PGD-50}  & $C\&W$  & \textit{AA}&\textit{CO}\\ 
		\hline  
		\multirow{4}{*}{{$\mathcal{M}_\text{A}$}} 
        &100&64.95&40.41&25.9&20.38&16.91&$\usym{2713}$\\
		&200&57.28&36.12&24.35&20.01&16.84 &$\usym{2713}$\\
        &400& 10.00&10.00&10.00&10.00&10.00&{\bf --}\\
        &1000& 10.00&10.00&10.00&10.00&10.00&{\bf --}\\
        \hline
		\multirow{4}{*}{{$\mathcal{M}_\text{B}$}} &100&60.18&37.91&24.18&20.11&16.77&$\usym{2713}$\\
		&200& {\bf 74.08}&{\bf 48.09}&29.65&25.32&19.72&$\usym{2717}$\\
        &400&72.62&47.74 &{\bf 29.76} &25.41 &19.85&$\usym{2717}$\\
        &1000& 71.30&46.89&29.73&{\bf 25.53}&{\bf 20.97}&$\usym{2717}$\\
        \hline
		\multirow{4}{*}{{$\mathcal{M}_\text{C}$}} &100
        &54.45&34.76&24.43&20.06&17.32&$\usym{2713}$\\
		&200& 58.11&37.57&25.69&20.49&17.16&$\usym{2713}$\\
        &400&63.66&38.11&24.30&18.71&15.59&$\usym{2713}$\\
        &1000&55.54&33.45&23.25&18.71&16.07&$\usym{2713}$\\
        \hline
		\multirow{4}{*}{{$\mathcal{M}_\text{D}$}} 
        &100&63.89&38.83&26.15&21.36&17.76&$\usym{2713}$\\
		&200& 58.76&36.43&25.30&19.84&17.17&$\usym{2713}$\\
        &400& 51.45&31.88&22.93&18.35&16.26&$\usym{2713}$\\
        &1000& 44.86&30.32&24.12&20.35&18.55&$\usym{2713}$\\
        \hline
		\multirow{4}{*}{{$\mathcal{M}_\text{E}$}} &100&49.72&32.83&24.03&22.47&19.87&$\usym{2713}$\\
		&200&54.55&34.97&25.68&22.39&19.68&$\usym{2713}$\\
        &400&55.20&36.67&26.48&22.67&19.94&$\usym{2713}$\\
        &1000& 10.00&10.00&10.00&10.00&10.00&{\bf --}\\
        \hline
	\end{tabular}
    \vspace{-6mm}
    \end{center}
	\caption{Mitigating CO by suppressing feature differences at various activation nodes. $\xi_T$ = 12/255 and $\xi_E$ = 16/255. {\bf --} means the failed convergence.}
	\label{tabsup}
 \vspace{-2mm}
\end{table}

Overall, the misclassification of CO-affected models on adversarial examples can be attributed to feature coverage rather than feature destruction. 
Namely, adversarial perturbations do not greatly affect the extracted data information. Instead, they deceive models by increasing feature values of several specific channels.
This explains why models can accurately recognize adversarial examples 
after removing the attack information based on Eq. (\ref{eq3}).

\subsection{Mitigating or inducing CO}
Sec. \ref{sec3.1} indicates that CO occurs with a significant increase of feature activation differences.
We further validate this phenomenon by manipulating these differences to mitigate or induce CO.
Specifically, to address CO, we introduce a supplemental constraint for Eq. (\ref{re1}), which can suppress feature activation differences,
\begin{equation}\label{act}
\centering
    \begin{aligned}
    \mathcal{L}_{stable} = \frac{\alpha_3}{\text{N}\cdot\text{H}\cdot\text{W}}\cdot\sum_k V_{act}^{{i},k}(x+\delta, x+\delta_0),
    \end{aligned}
\end{equation}
where $k$ denotes the $k$-th activation channel of the $i$-th activation node. 
$\alpha_3$ can be assigned a large value since $V_{act}^{{i},k}(x+\delta, x+\delta_0)$ converges towards zero under stable adversarial training.
$\delta_0$ and $\delta$ indicate the uniformly random perturbations and crafted adversarial perturbations, respectively. 
$\mathcal{L}_{stable}$ establishes consistency between $\mathcal{M}_{i,k}(x+\delta)$ and $\mathcal{M}_{i,k}(x+\delta_0)$ rather than $\mathcal{M}_{i,k}(x)$. This reduces training time as $\mathcal{M}_{i,k}(x+\delta_0)$ is already calculated when generating adversarial perturbations $\delta$. 

\begin{table}
	\centering
	\tabcolsep = 0.085cm
 \begin{center}
	\begin{tabular}{ccccc cccc} 
		\hline
		 \multicolumn{2}{c}{Res-CIFAR10} &$\xi_T$    &  \textit{Clean}& \textit{PGD-50}  & $C\&W$  & \textit{AA}& \textit{Mins} \\ 
		\hline  
  \multicolumn{2}{c}{\textit{FGSM-MEP} \cite{Jiag2022prior}}& \multirow{7}{*}{$\frac{12}{255}$}& 74.71 &35.52 &33.05 &27.23& 92\\
		\multicolumn{2}{c}{\textit{C.Smooth} \cite{zhao2023fast}}& &71.79&27.95&24.12&19.11&104\\
        \cline{1-2}
		\multirow{5}{*}{\textit{Ours}} 
        &$\alpha_3$=100& &60.18&24.18&20.11&16.77&{ 93}\\
		&$\alpha_3$=200& & {\bf 74.08}&29.65&25.32&19.72&{ 93}\\
        &$\alpha_3$=300& & 73.13&29.36&25.06&19.58&{ 93}\\
        &$\alpha_3$=400& &72.62&{\bf 29.76} &25.41 &19.85&{ 93}\\
        &$\alpha_3$=1000& & 71.30&29.73&{\bf 25.53}&{\bf 20.97}&{ 93}\\
        \hline
      \multicolumn{2}{c}{\textit{FGSM-MEP} \cite{Jiag2022prior}}& \multirow{3}{*}{$\frac{16}{255}$}&53.32 &26.56 &22.10 &18.98& 92\\
        \multicolumn{2}{c}{\textit{C.Smooth} \cite{zhao2023fast}}&  &63.84 &32.95 &28.19 &{\bf 23.68} &104\\
        \textit{Ours} &$\alpha_3$=200 &&{\bf 65.01}&{\bf 34.17}&{\bf 29.79}&22.85&{\bf 93}\\
        \hline\hline
		 \multicolumn{2}{c}{Res-CIFAR100} &$\xi_T$    &  \textit{Clean}& \textit{PGD-50}  & $C\&W$  & \textit{AA}& \textit{Mins} \\
   \hline

   		\multicolumn{2}{c}{\textit{C.Smooth} \cite{zhao2023fast}}& \multirow{6}{*}{$\frac{12}{255}$} &50.61&14.70&12.54&9.04&104\\
     \cline{1-2}
     \multirow{5}{*}{\textit{Ours}} 
        &$\alpha_3$=100& &{\bf 51.27}&14.81&12.71&9.82&{ 93}\\
		&$\alpha_3$=200& &51.01&14.86&{\bf 13.06}&{\bf 10.08} &{ 93}\\
        &$\alpha_3$=300& &50.51&{\bf 14.87}&12.69&9.83 & { 93}\\
        &$\alpha_3$=400& &50.44&14.49&12.41&9.64&{ 93}\\
        &$\alpha_3$=1000& &50.04&14.52&12.29&9.67&{ 93}\\
        \hline
        \multicolumn{2}{c}{\textit{C.Smooth} \cite{zhao2023fast}}& \multirow{2}{*}{$\frac{16}{255}$} &41.86&{\bf 16.59}&{13.96}&{\bf 11.38}&{104}\\
        \textit{Ours} &$\alpha_3$=200 &&{\bf 42.68}&{16.47}&{\bf 14.40}&11.01&{ 93}\\
        \hline
	\end{tabular}
 \end{center}
 \vspace{-4mm}
 \caption{Comparison with the state-of-the-art methods. `Res' denotes the network ResNet18.
 The models are trained with perturbation budgets of 12/255 and 16/255, and evaluated with the perturbation budget of 16/255.
 `\textit{Mins}' is the minutes for training. Only the node $\mathcal{M}_\text{B}$ is used to calculate $\mathcal{L}_{stable}$ in Eq. (\ref{act}).}
	\label{tab3}
 \vspace{-2mm}
\end{table}

The results in Tab. \ref{tabsup} show that only suppressing the feature activation differences at $\mathcal{M}_\text{B}$ can resolve CO. In our opinion, this is due to $\mathcal{M}_\text{B}$ being more sensitive to CO, as shown in Fig. \ref{fig1}.
Tabs. \ref{tab3} and \ref{tab2} demonstrate that our approach achieves better or comparable performance than the state-of-the-art methods. 
Particularly, using the same hyperparameter ($\alpha_3$ = 200) can effectively mitigate CO across diverse datasets and networks.
This alleviates the burden of hyperparameter selection in \cite{zhao2023fast}.
More detailed experimental analyses are given in supplementary materials.

\begin{figure}[t]
    \begin{center}
        \includegraphics[width=1\linewidth]{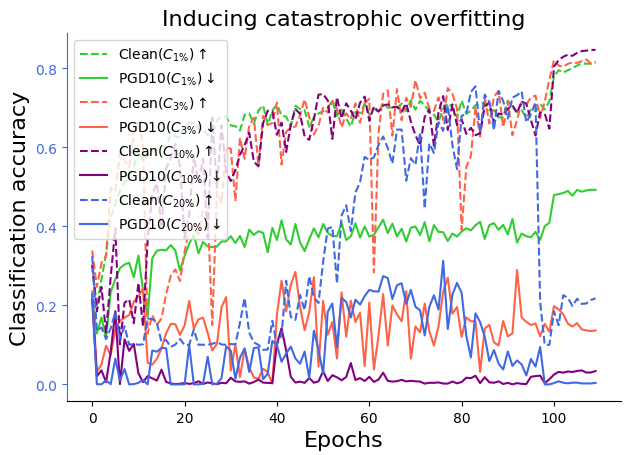}
    \end{center}
    \vspace{-6mm}
    \caption{Investigate whether $\mathcal{L}_{co}$ in Eq. (\ref{eq4}) can induce CO.}
    \label{fig12}
    \vspace{-2mm}
\end{figure}

\begin{figure}[tpb]
    \begin{center}
        \includegraphics[width=1\linewidth]{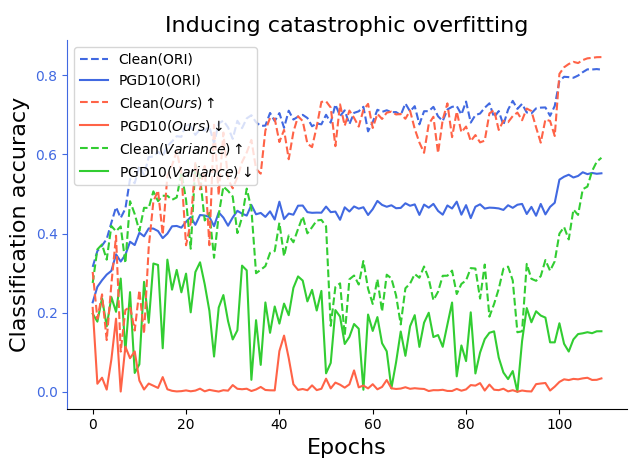}
    \end{center}
    \vspace{-6mm}
    \caption{Comparison with prior work \cite{he2023investigating} in inducing CO.}
    \label{fig12_2}
    \vspace{-2mm}
\end{figure}

Next, we explore how to induce CO.
$\xi_T$ is set to 8/255 since the adversarial training process is stable under this setting.
$\xi_E$ = 16/255.
Sec. \ref{sec3.2} shows that only several specific channels are responsible for CO.
Thus, we first select the feature channels with the top $p\%$ largest $T_{act}^{\text{B},k}$ (denoted as $C_{p\%}$) based on the statistical information of training dataset.
Then, we design the supplemental constraint for Eq. (\ref{re1}),
\begin{equation}\label{eq4}
\centering
    \begin{aligned}
    \mathcal{L}_{co} = \mathcal{L}\left(\mathcal{M}(x+\delta, C_{p\%}), y\right) - \mathcal{L}_{stable}(C_{p\%}),
    \end{aligned}
\end{equation}
where $\mathcal{M}(\cdot, C_{p\%})$ 
denotes the models masked $C_{p\%}$.
The former item ensures that the masked models can correctly classify adversarial examples, and the latter enlarges the feature differences of channels $k \in C_{p\%}$. 
$p$ is set to 1, 3, 10, and 20 respectively.
As depicted in Fig. \ref{fig12}, $\mathcal{L}_{co}$ with $C_{10\%}$ can bring about CO.  Moreover, adding $\mathcal{L}_{co}$ to $C_{1\%}$ is insufficient to induce CO, whereas introducing $\mathcal{L}_{co}$ to $C_{20\%}$ prevents the model from converging. Furthermore, we conduct a comparative experiment based on \cite{he2023investigating} by selecting feature channels with the top $10\%$ maximum channel variance. Fig. \ref{fig12_2} shows that the model is difficult to converge under this setting. In our opinion, this is because several selected channels mainly capture data information, and increasing their variance affects the model optimization.


\section{Leveraging CO to enhance performance}
\begin{figure}[t]
    \begin{center}
        \includegraphics[width=1\linewidth]{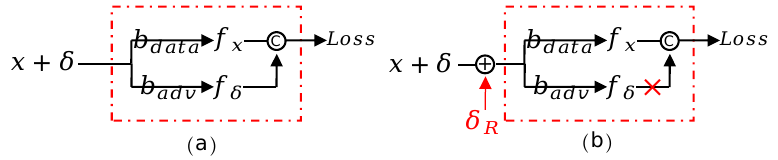}
    \end{center}
    \vspace{-6mm}
    \caption{ The structure of models that suffer from CO.
    (a) Original structure; (b) The structure of adding random noise $\delta_R$. $f$ denotes the extracted features.}
\label{fig5}
\vspace{-2mm}
\end{figure}

In the preceding section, we have discussed the underlying causes of CO and its potential solutions. 
Drawing from these insights, we then illustrate whether CO has to be solved. 
Specifically,
we try to improve the performance of CO-affected models by leveraging their different pathways for extracting data and attack information.
Although these pathways are not entirely separable, we describe them as two distinct branches in Fig. \ref{fig5} (a) for clarity, \ie, the data branch `$b_{data}$' and the adversarial branch `$b_{adv}$'.


\subsection{Attack obfuscation} 
We expect that adversarial attacks mainly deceive the branch $b_{adv}$ and that the models primarily use features extracted from the branch $b_{data}$ to classify both clean and adversarial examples,
which is called attack obfuscation.
Fortunately, models that suffer from CO inherently satisfy the former expectation,
as shown in Fig. \ref{fig3} and Fig. \ref{fig4}.
The latter contains twofold: 1) the model performance in regular tasks should be preserved; 2) the attack information carried by adversarial perturbations can be disrupted.
These can be realized by introducing random noise $\delta_R$ to the input of CO-affected models during inference, as shown in Fig. \ref{fig5} (b).
To explain this, we take the uniformly random noise ($\delta_R\sim \text{U}(-16/255,16/255)$) as an example.
In the following, $\delta_0$, $\delta$, and $\delta_{R}$ represent the initial adversarial perturbation, the crafted adversarial perturbation, 
and the random noise added to model input, respectively.

\begin{table}[t]
	\tabcolsep = 0.1cm
 \begin{center}
	\begin{tabular}{cccc} 
		\hline 
		&\textit{Origin}&\textit{CO.}&\textit{Stable}\\ 
        \hline
        \textit{w/o.} $\delta_R$&94.40&89.81&72.84\\
        \textit{w.} $\delta_R$ &80.63&89.64&72.06\\
        \hline
	\end{tabular}
 \vspace{-4mm}
 \end{center}
	\caption{Classification accuracy of various models on clean samples.
 `\textit{Origin}' denotes the performance of the original ResNet18.
 `\textit{Stable}' represents the stably trained model.
 `\textit{CO.}' means the model suffering from catastrophic overfitting.
 $\xi_T$ = 12/255 and $\delta_R\sim$ \text{U}(-16/255,16/255).
 \text{U}($\cdot$,$\cdot$) is uniformly random noise.
 }
	\label{tab1}
 \vspace{-3mm}
\end{table}

\begin{figure}[t]
    \begin{center}
        \includegraphics[width=1.\linewidth]{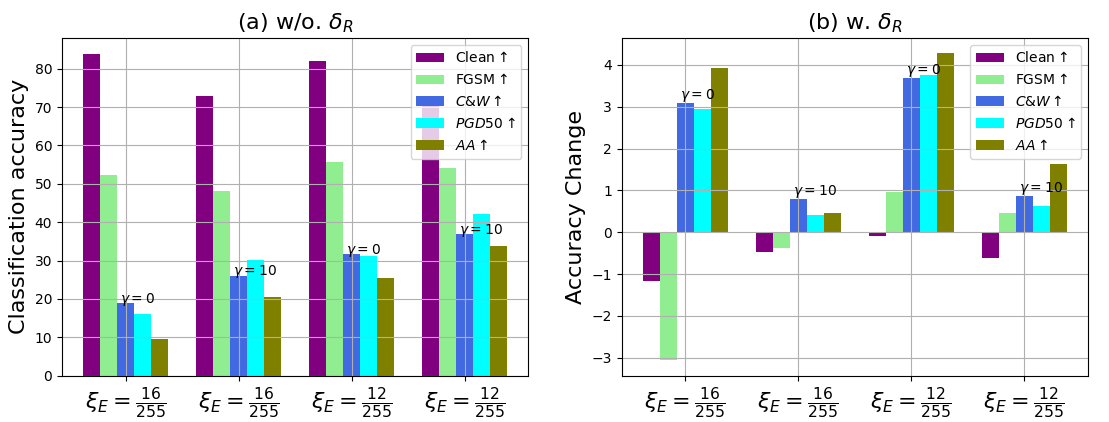}
    \end{center}
    \vspace{-4mm}
    \caption{
Relationship between adversarial robustness and capacity of withstanding noise interference for adversarial perturbations.
$\xi_T$ = 12/255. $\xi_E$ is set to 12/255 or 16/255.
(a) Classification accuracy of models.
(b) 
The degree of change in classification accuracy when adding uniformly random noise $\delta_R\sim$U(-16/255,16/255).
    }
    \label{fig6}
    \vspace{-2mm}
\end{figure}

\begin{figure}[htpb]
    \begin{center}
    \vspace{-2mm}
        \includegraphics[width=0.9\linewidth]{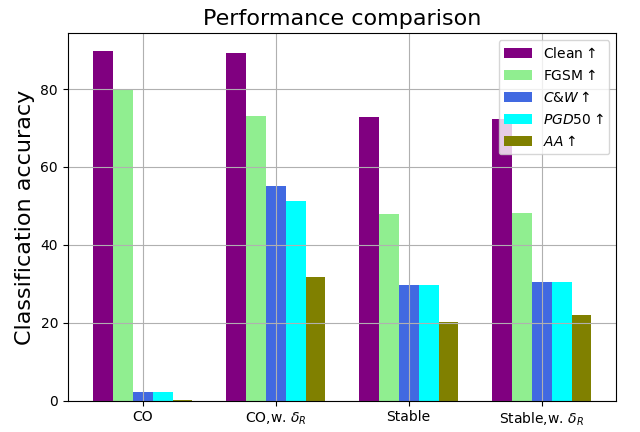}
    \end{center}
    \vspace{-6mm}
    \caption{Classification accuracy of various models on clean and adversarial data.
    $\xi_T$ = 12/255, $\xi_E$ = 16/255, and $\delta_R$ $\sim$ U(-16/255,16/255).
    \text{U}($\cdot$,$\cdot$) is uniformly random noise.
    }
    \vspace{-4mm}
    \label{fig7}
\end{figure}

It can be observed in Tab. \ref{tab1} that the model suffering from CO displays well generalization to $\delta_R$.
Meanwhile, we find in Fig. \ref{fig7} that adversarial perturbations for deceiving CO-affected models maintain the limited attack ability after adding $\delta_R$.
This is a common phenomenon because:
1) Compared to stably trained models, CO-affected models show a significant decrease in adversarial robustness;
2) As the adversarial robustness of victim models decreases, the capacity of withstanding noise interference weakens for adversarial perturbations.
To validate this, we train two comparative models with $\xi_T$ = 12/255. The distinction between them is whether using the regularization term in \cite{Jiag2022prior},
\begin{equation}\label{eq5}
\centering
    \begin{aligned}
\gamma\cdot||\mathcal{M}(x+\delta) - \mathcal{M}(x + \delta_{0})||_2^2.
    \end{aligned}
\end{equation}
The adversarial examples for evaluation are generated under two perturbation budgets: 12/255 and 16/255. 
Fig. \ref{fig6} (a) shows that Eq. (\ref{eq5}) can greatly improve the adversarial robustness of models.
Furthermore, Fig. \ref{fig6} (b) indicates that adding random noise $\delta_R$ has the limited influence on the models with strong adversarial robustness.

\begin{table}[htpb]
	\centering
	\tabcolsep = 0.08cm
 \begin{center}
	\begin{tabular}{ccc cccc} 
		\hline
        \textit{U(-a, a)}& \textit{Gauss ($\sigma$)} & \textit{Clean}&\textit{PGD-50} & \textit{C$\&$W} & \textit{AA} \\
        \hline
        a = 10/255&-&88.26&33.85&38.90&23.27\\
        a = 12/255&-& 88.44&46.49&50.01&33.70\\
        a = 14/255&-& 88.39& 54.07& 58.03&40.56\\
        a = 16/255&-& {\bf 88.45}& {\bf 58.48}& {\bf 61.65}& {\bf 43.53}\\
        \hline
        -& 10/255&87.80&54.25&58.14&39.73\\
        -&12/255&86.41&56.74&60.22&41.64\\
        -&14/255&83.84&57.26&59.74&41.48\\
        -&16/255&80.25&55.81&58.68&40.34\\
        \hline
	\end{tabular}
 \end{center}
	\vspace{-4mm}
 \caption{
 Ablation studies for types of random noise $\delta_R$ on the robustness of CO-affected models.
Experiments are conducted on CIFAR10 with ResNet18. 
U($\cdot$, $\cdot$) is uniformly random noise, while \textit{Gauss} denotes Gaussian noise with an average of 0. $\sigma$ is the standard deviation of $\textit{Gauss}$.
 }
	\label{tabz4}
 \vspace{-4mm}
\end{table}

\begin{table*}
	\centering
	\tabcolsep = 0.13cm
 \begin{center}
 \vspace{-2mm}
	\begin{tabular}{ccccc cccccc} 
		\hline
		\textit{$\xi_T$}& \textit{Methods}     &  \textit{Clean$\uparrow$}  &\textit{FGSM$\uparrow$} & \textit{PGD-10$\uparrow$}  &  \textit{PGD-20$\uparrow$} & \textit{PGD-50$\uparrow$}  & \textit{C$\&$W$\uparrow$}  & \textit{APGD-T\cite{Francesco2020}$\uparrow$} & \textit{AA $\uparrow$}\\ 
		\hline  
        \multirow{5}{*}{\textit{10/255}}
		&\textit{ATAS} \cite{huang2022fast}&83.43&48.09&31.29&19.43&15.37&18.55&12.93&10.94\\
		&\textit{ZeroGrad} \cite{golgooni2023zerograd} &82.70&45.32&30.26&18.59&14.93&17.59&12.47&10.97\\
		&\textit{C.Smooth} \cite{zhao2023fast}&76.10&47.12&37.30&28.16&26.11&23.48&19.32&18.23\\
        &\textit{Ours-FD.}&76.73&48.85&39.59&30.82&28.61&26.58&21.91&20.07\\
        &\textit{Ours-CO.} &88.91 &70.07 &58.45 &48.04 &38.63 &49.41 &41.31 &19.45\\
        \hline  
        \multirow{5}{*}{\textit{12/255}}
		&\textit{ATAS} \cite{huang2022fast} &83.52&49.02&33.25&21.78&17.73&21.15&14.06&   11.91\\
		&\textit{ZeroGrad} \cite{golgooni2023zerograd}&82.17&48.98&33.81&22.37& 18.34&21.65&14.98&12.85\\
		&\textit{C.Smooth} \cite{zhao2023fast}& 71.79&45.73&37.79&29.81&27.95&24.12&19.71&19.11\\
        &\textit{Ours-FD.} &72.60&49.08&40.87&33.06&31.42&28.16&23.33&22.08\\
        &\textit{Ours-CO.}&{\bf 89.39} &{\bf 73.18} &64.15 &57.32 &51.25 &58.82 &55.13 &31.62\\
        \hline
        \multirow{5}{*}{\textit{14/255}}
		&\textit{ATAS} \cite{huang2022fast}&81.54&48.96&34.24&23.05&19.54&21.91&15.69&13.38 \\
		&\textit{ZeroGrad} \cite{golgooni2023zerograd}&74.23&46.76&35.42&25.27& 21.94&23.19&18.03&15.64\\
		&\textit{C.Smooth} \cite{zhao2023fast}&65.57&45.40&38.75&32.33&31.20&25.97&23.15&22.51\\
        &\textit{Ours-FD.} &68.98&47.67&40.55&33.96&32.56&28.10&24.36&23.54\\
        &\textit{Ours-CO.}&89.12 &71.48 &61.46 &53.72 &46.29 &54.29 &56.01 &30.65\\
        \hline
        \multirow{5}{*}{\textit{16/255}}
		&\textit{ATAS} \cite{huang2022fast}&79.91&52.28&38.96&28.64&25.16&27.83&23.82&16.95\\
		&\textit{ZeroGrad} \cite{golgooni2023zerograd}&67.34 &42.07&32.46&25.35 &22.45 &21.33 &15.79 &15.15\\
		&\textit{C.Smooth} 
         \cite{zhao2023fast}&62.88&44.98&40.98&35.48&34.29&29.94&26.23&25.10\\
        &\textit{Ours-FD.}&64.08&48.02&42.56&37.43&36.83&32.51&28.62&27.37\\
        &\textit{Ours-CO.}&88.45 &71.59 &{\bf 66.27} &{\bf 60.59} &{\bf 58.48} &{\bf 61.65} &{\bf 62.03} &{\bf 43.53}\\
        \hline
	\end{tabular}
 \end{center}
 \vspace{-6mm}
	\caption{Quantitative results of various FAT methods when adding $\delta_R\sim$U(-16/255,16/255) to inputs. We adopt ResNet18 and CIFAR-10. U($\cdot$,$\cdot$) is uniformly random noise. Models are evaluated with $\xi_E$ = 16/255. Our methods employ the same baseline as \cite{zhao2023fast}. For ATAS and ZeroGrad, we employ the early stopping technique \cite{sitawarin2021sat} when these methods falling into CO. \textit{FD} means the proposed feature difference.}
	\label{tab2}
\vspace{-2mm}
\end{table*}

{\bf Performance comparison of various models.}
In Tab. \ref{tab1}, CO-affected models exhibit higher classification accuracy on clean samples (89.81\%) than stably trained models (72.84\%).
Furthermore, they display excellent generalization performance to noise-perturbed examples, achieving an accuracy of 89.64\%.
Additionally, experimental results in Fig. \ref{fig7} and Tab. \ref{tab2} show that adding uniformly random noise to the input of CO-affected models can attain the optimal adversarial robustness.
In summary, CO can be utilized to enhance the model performance on both clean and adversarial examples.
Experiments on various datasets and networks in supplementary materials also support this conclusion.

\begin{figure*}[t]
   \vspace{-2mm}
    \begin{center}
        \includegraphics[width=1\linewidth]{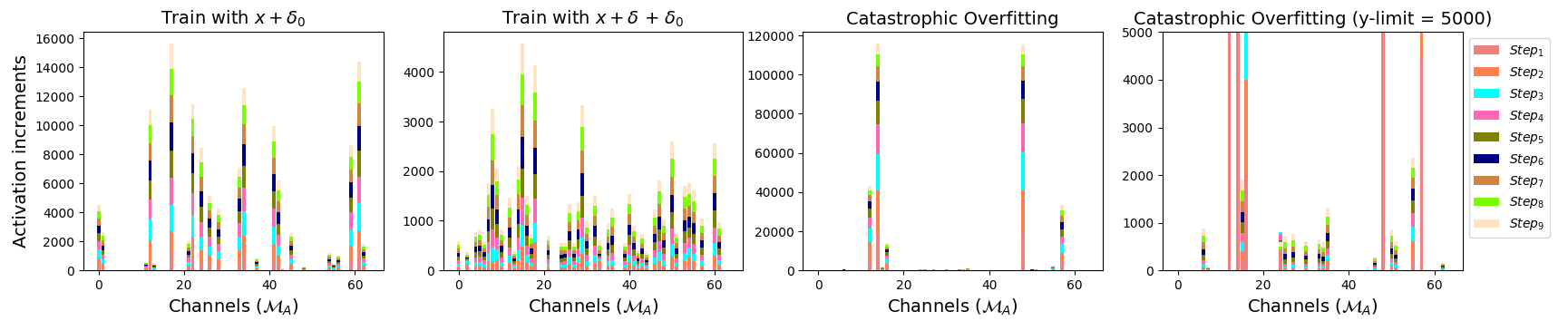}
    \end{center}
    \vspace{-6mm}
    \caption{Activation increments observed in various models at each step of the PGD-10 attack. We statistic the results within the test set. $Step_i$ is activation increments induced by the $i$-th adversarial attack. $\delta$ and $\delta_0$ denote adversarial and random perturbations, respectively.}
    \label{fig9}
    \vspace{-4mm}
\end{figure*}

{\bf Ablation studies on types of random noise $\delta_R$.}
In addition to applying uniformly random noise, 
we also explore the impact of Gaussian noise on the performance of CO-affected models ($\xi_T$ = 16/255). 
For Gaussian noise, we vary the standard deviation with values of 10/255, 12/255, 14/255, and 16/255, keeping the mean value at 0. The experimental results are shown in Tab. \ref{tabz4}. It can be observed that adding $\delta_R$ $\sim$ U(-16/255,16/255) to model inputs achieves the optimal classification accuracy on both clean and adversarial data. More results about models under various $\xi_T$ can be found in supplementary materials.

\subsection{Necessity of inducing CO for robustness}\label{sec4.2}
\begin{figure}[t]
    \begin{center}
    \vspace{-2mm}
        \includegraphics[width=1.\linewidth]{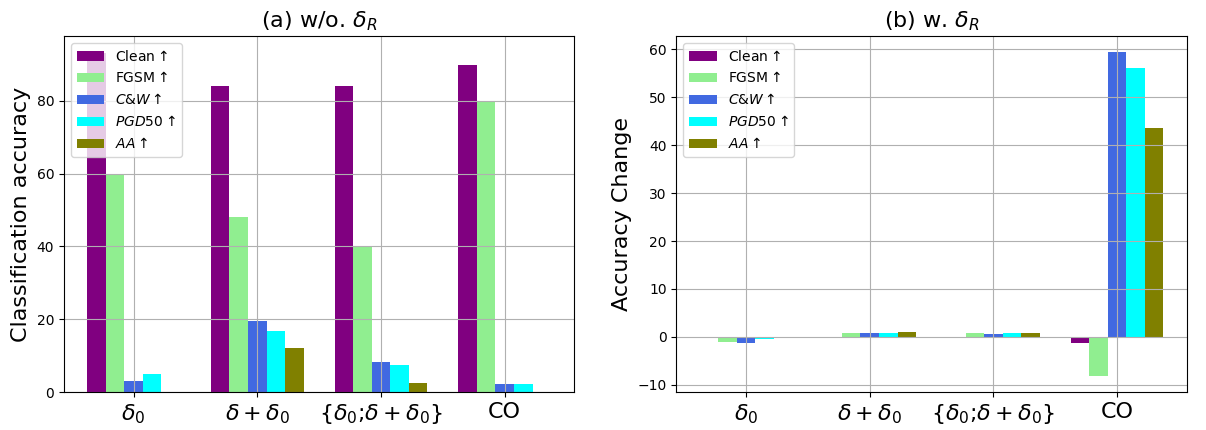}
    \end{center}
    \vspace{-6mm}
    \caption{
    Comparative experiments on models trained with various noise-augmented data. (a) Classification accuracy of initial models. (b) The degree of change in classification accuracy when adding $\delta_R\sim$U(-16/255,16/255) to model inputs. $\delta_0$: augment training data by adding random noise. $\delta+\delta_0$: superposing adversarial perturbation and random noise. $\{\delta;\delta+\delta_0\}$: combining the noise-augmented data $x+\delta_0$ and $x+\delta+\delta_0$.
    }
\label{fig8}
\vspace{-4mm}
\end{figure}

We believe that the robustness improvement of CO-affected models is not solely ascribed to their generalization ability to noise.
To verify this, we conduct comparative experiments by utilizing various noise-augmented data to train models.
We first augment training data by adding uniformly random noise $\delta_0$, with $\xi_T$ = 16/255,
\begin{equation}\label{eq6}
    \min_\theta \mathbb{E}_{(x, y)\sim D_{train}} [\mathcal{L}(\mathcal{M}(x+\delta_0), y)].
\end{equation}
As depicted in Fig. \ref{fig8} (a), the effective generalization to $\delta_0$ does not improve adversarial robustness.

We then assess the augmentation strategy of superposing adversarial perturbation $\delta$ and uniformly random noise $\delta_0$. 
\begin{equation}\label{eq7}
    \min_\theta \mathbb{E}_{(x, y)\sim D_{train}}[ \mathcal{L}(\mathcal{M}(x+\delta+\delta_0), y) ].
\end{equation}
In Fig. \ref{fig8} (a), the generalization to $\delta+\delta_0$ helps improve the adversarial robustness.
However, this benefit is considerably lower than that of adding $\delta_R$ to the input of CO-affected models, as shown in Fig. \ref{fig8} (b).
It is also evident that utilizing both $x+\delta_0$ and $x+\delta+\delta_0$ as training data does not result in stronger adversarial robustness.

\begin{figure*}[t]
    \begin{center}
    \vspace{-2mm}
        \includegraphics[width=0.95\linewidth]{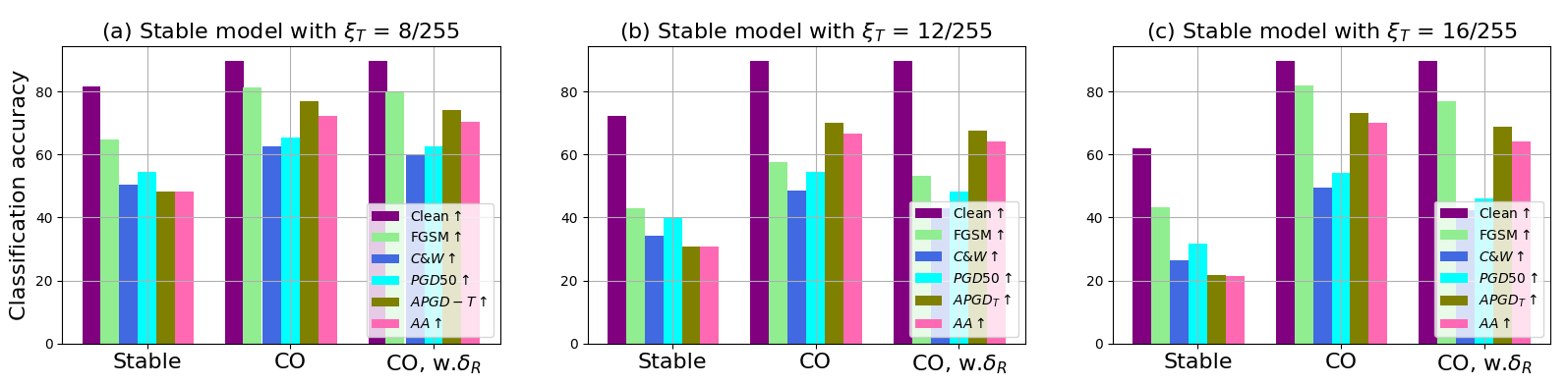}
    \end{center}
    \vspace{-7mm}
    \caption{Classification accuracy of CO-affected models ($\xi_T$ = 12/255) for transferred adversarial examples. Experiments are studied on CIFAR-10 with ResNet18.
    (a) $\xi_E$ = 8/255; (b) $\xi_E$ = 12/255; (c) $\xi_E$ = 16/255. $\delta_R$ $\sim$ U(-16/255,16/255). U($\cdot$, $\cdot$) is uniformly random noise.}
    \label{fig11}
    \vspace{-3mm}
\end{figure*}

To understand the differences between CO-affected and noise-augmented models,
we display channel-specific activation increments $\sum_{x\in D_{test}}\frac{V_{act}^{{i},k}(x, x+\delta_j)}{\text{N}\cdot \text{H}\cdot \text{W}}$ at each step $j$ of adversarial attacks in Fig. \ref{fig9}.
Comparatively, CO-affected models exhibit salient activation increments in several specific channels, 
while noise-augmented models show comparable activation increments across different channels.
Hence, we think that for noise-augmented models, the channels responsible for learning data and attack information are nearly inseparable. 
Consequently, adversarial attacks on these models are more likely to affect the channels of learning data information.
For instance, many activation increments in noise-augmented models surpass $1000$, exceeding the levels observed in CO-affected models.
These findings, coupled with the performance comparison between CO-affected and noise-augmented models in Fig. \ref{fig8}, indirectly validate the fact of `attack obfuscation'.

Crucially, for the models trained on clean samples, adding uniformly random noise to their input does not improve the adversarial robustness. 
The experimental results are shown in Tab. \ref{tabsup2}.
It can be seen that these models are not robust against adversarial attacks such as PGD and C$\&$W.

%


\subsection{Robustness against transferred $x+\delta$}
We explore the performance of CO-affected models when they are exposed to transferred adversarial examples. Specifically, we initially train stable models using \cite{zhao2023fast} under perturbation budgets of 8/255, 12/255, and 16/255, respectively. Subsequently, we generate adversarial examples against these stable models and evaluate the attack performance of these examples on a CO-affected model ($\xi_T$ = 12/255).
The results on CIFAR10 are provided in Fig. \ref{fig11}, which demonstrate the weak attack performance of these adversarial examples.
Additionally, we find a similar phenomenon on CIFAR100 in supplementary materials.

\subsection{Various FAT baselines}
In previous sections, we primarily utilize the FGSM-MEP as the FAT baseline. Here, we conduct experiments based on FGSM-RS \cite{Wong2020}.
It can be observed from Tab. \ref{tabsup3} that applying random noise $\delta_R$ to the model trained by FGSM-RS also significantly improves its adversarial robustness.

\begin{table}[t]
	\tabcolsep = 0.1cm
 \begin{center}
	\begin{tabular}{cccccc} 
		\hline 
		Res-CIFAR10&\textit{Clean}&\textit{FGSM}&\textit{PGD-50}&\textit{C\&W}&\textit{AA}\\ 
        \hline
        \textit{w/o.}$\delta_R$&94.40& 2.13&0.00&0.00&0.00\\
        \textit{w.}$\delta_R$&80.63&4.07&0.00&0.00&0.00\\
        \hline\hline
        Res-CIFAR100&\textit{Clean}&\textit{FGSM}&\textit{PGD-50}&\textit{C\&W}&\textit{AA}\\ 
        \hline
        \textit{w/o.}$\delta_R$&76.64&1.31&0.00&0.00&0.00\\
        \textit{w.}$\delta_R$&47.53&3.02&0.00&0.00&0.00\\
        \hline
	\end{tabular}
 \vspace{-4mm}
 \end{center}
	\caption{Classification accuracy of models trained on clean samples. `Res’ denotes the network ResNet18. 
 $\delta_R$ $\sim$ U(-16/255,16/255).
 U($\cdot$,$\cdot$) denotes the uniformly random noise.}
	\label{tabsup2}
 \vspace{-4mm}
\end{table}
\begin{table}[t]
	\tabcolsep = 0.12cm
 \begin{center}
	\begin{tabular}{cccccc} 
		\hline 
		$\xi_T$&$\delta_R\sim$U(-a,a)&\textit{Clean}&\textit{PGD-50}&\textit{C\&W}&\textit{AA}\\ 
        \hline
        \multirow{2}{*}{10/255}&a = 0.&79.97&2.66&4.15&0.00\\
        &a = 16/255&85.84&38.99&44.00&25.85\\
        \hline
        \multirow{2}{*}{12/255}&a = 0.&85.10&0.50&0.51&0.00\\
        &a = 16/255&86.58&39.25&45.72&26.03\\
        \hline
        \multirow{2}{*}{14/255}&a = 0.&82.81&0.14&0.39&0.00\\
        &a = 16/255&87.59&44.99&52.76&24.55\\
        \hline
        \multirow{2}{*}{16/255}&a = 0.&70.97&0.16&0.29&0.00\\
        &a = 16/255&85.75&46.55&53.29&26.89\\
        \hline
	\end{tabular}
 \vspace{-6mm}
 \end{center}
	\caption{
 Classification accuracy of models trained by \cite{Wong2020}.
We adopt the dataset CIFAR10 and the network ResNet18. U($\cdot$,$\cdot$) denotes the uniformly random noise.}
	\label{tabsup3}
 \vspace{-4mm}
\end{table}

\section{Conclusion}
In this work, we provide a comprehensive explanation for CO by analyzing the feature activation differences between clean and adversarial examples. 
Our studies reveal that CO occurs with salient feature activation differences.
On this basis, we propose novel regularization terms to either mitigate or induce CO by manipulating feature differences, which indirectly validates the relationship between CO and these differences.
Importantly, our regularization terms are insensitive to hyperparameters. 
Furthermore, 
we incorporate random noise into CO-affected models, thereby achieving optimal classification accuracy for both clean and adversarial examples. 
To explain this phenomenon, we approach it from the perspective of attack obfuscation.

{
    \small
    \bibliographystyle{ieeenat_fullname}
    \bibliography{main}
}


\end{document}